\documentclass[runningheads]{llncs}

\usepackage{eccv}

\usepackage{eccvabbrv}

\usepackage{graphicx}
\usepackage{booktabs}
\usepackage{color}
\usepackage{tablefootnote}
\usepackage[accsupp]{axessibility} 

\usepackage{hyperref}

\usepackage{orcidlink}
\usepackage{dsfont}

\begin{document}

\title{Introducing Gating and Context into Temporal Action Detection} 


\author{Aglind Reka\inst{1}\orcidlink{0009-0003-6841-2984} \and
Diana Laura Borza\inst{3}\orcidlink{0000-0002-0276-7243} \and
Dominick Reilly\inst{2}\orcidlink{0009-0005-3959-8032} \and
Michal Balazia\inst{1}\orcidlink{0000-0001-7153-9984}\and
Francois Bremond\inst{1}\orcidlink{0000-0003-2988-2142}}

\authorrunning{A. Reka, D. Borza}

\institute{Inria Sophia Antipolis, France \\
\email{\{aglind.reka,michal.balazia,francois.bremond\}@inria.fr} \and
University of North Carolina at Charlotte \\
\email{dreilly1@charlotte.edu}\and
Babes Bolyai University, Romania \\
\email{diana.borza@ubbcluj.ro}}

\maketitle


\begin{abstract}
Temporal Action Detection~(TAD), the task of localizing and classifying actions in untrimmed video, remains challenging due to action overlaps and variable action durations. Recent findings suggest that TAD performance is dependent on the structural design of transformers rather than on the self-attention mechanism. Building on this insight, we propose a refined feature extraction process through lightweight, yet effective operations. First, we employ a local branch that employs parallel convolutions with varying window sizes to capture both fine-grained and coarse-grained temporal features. This branch incorporates a gating mechanism to select the most relevant features. Second, we introduce a context branch that uses boundary frames as key-value pairs to analyze their relationship with the central frame through cross-attention. The proposed method captures temporal dependencies and improves contextual understanding. Evaluations of the gating mechanism and context branch on challenging datasets~(THUMOS14 and EPIC-KITCHEN 100) show a consistent improvement over the baseline and existing methods.
\end{abstract}

\section{Introduction}
\label{sec:intro}

Temporal Action Detection~(TAD) involves localizing actions within untrimmed video sequences. Despite significant advances in the field, several challenges remain unaddressed. Accurate action detection requires capturing temporal relationships between frames, a challenging task because of simultaneous actions and varying duration of actions. Moreover, distinguishing between similar actions also requires understanding the context in which the action occurs, and feature extraction without inducing excessive complexity remains a problem in long, untrimmed videos.

Recent work of Shi~\emph{et al.}~\cite{shi2023tridet} has demonstrated that the performance of TAD models is not necessarily dependent on the self-attention mechanism but rather on the macro architecture of the transformer. Inspired by this finding, our study builds upon the TriDet architecture~\cite{shi2023tridet} and focuses on refining the feature extraction process.

The contributions of this work are two-fold. The proposed method relies on a local branch that employs two parallel convolutions with different window sizes to capture both fine-grained and coarse-grained temporal features. We introduce a gating mechanism to select the most discriminatory features based on the convolution output. This selective process increases the model's ability to prioritize relevant information in action detection. In addition, We introduce a context branch that utilizes peripheral frames in the receptive field of the convolutional kernel as key-value pairs and examines their relationship with the central frame through cross-attention. This approach captures temporal dependencies, providing a more nuanced contextual understanding of action sequences. Finally, we perform experiments and ablation studies on challenging datasets~(THUMOS14~\cite{idrees2017thumos} and EPICKITCHEN 100~\cite{damen2022rescaling}) to assess the impact of the gating mechanism and the context branch.

\section{Related Work}
\label{sec:related_work}
This section discusses recent advancements and approaches in TAD, including both two-stage and one-stage models, as well as the importance of feature selection and gating mechanisms in improving deep learning models' performance.

\paragraph{Two stage TAD.}
From a high-level perspective, TAD models can be categorized into one-stage and two-stage models.
Two-stage methods~\cite{buch2017sst,lin2020fast,zhu2021enriching,bai2020boundary}, similar to two-stage object detectors, approach the action localization problem by dividing it into two distinct stages: proposal generation and subsequent classification of these proposals. The main focus of these works is to generate action proposals, either by classifying actions from specific anchor windows~\cite{buch2017sst,lin2020fast} or by predicting the action boundaries~\cite{zhu2021enriching,bai2020boundary}. The main drawbacks of these methods are their inherent complexity and their inability to be trained in an end-to-end manner.

\paragraph{One stage TAD.}
On the other hand, one-stage methods aim to identify actions in videos in a single step, without the preliminary action proposals. These methods streamline the action detection process, making it faster and more efficient. Based on the success of transformer models in the field of computer vision, the majority of TAD models rely on attention mechanisms. PDAN~\cite{dai2021pdan} uses the Dilated Attention Layer~(DAL) as a building block to capture the local context through dilated convolutions and attention mechanisms. These layers are organized in a feature pyramid to handle actions with different temporal lengths. ActionFormer~\cite{zhang2022actionformer} employs a transformer-based architecture and constructs hierarchical representations of video sequences to model and capture actions occurring at different temporal scales. Other works~\cite{shi2022react,liu2022end} draw inspiration from transformer-based object detection methods~\cite{carion2020end} and use a set of learnable input queries to decode a set of action predictions. TALLFormer~\cite{cheng2022tallformer} leverages a long-term memory module for action understanding. A short-term transformer module extracts features from a randomly sampled set of frames, while the other features are obtained from long-term memory. 

More recently, TriDet~\cite{shi2023tridet} demonstrated that the efficiency of recent TAD models is more related to the overall architecture of the transformer models than to the self-attention mechanism. The authors proposed the Scalable-Granularity Perception~(SGP) layer for feature extraction which uses solely convolution operations. This layer mitigates the rank loss problem across the temporal dimension and the high computational overhead of the attention mechanism. Additionally, the Tridet architecture models action boundaries by estimating a relative probability distribution around those boundaries. \cite{shi2023tridetplus} extends TriDet by building two separate feature pyramids using different backbones: the temporal-level feature pyramid and the spatial-level feature pyramid. The temporal-level feature pyramid is directly fed into the classification head. Meanwhile, both pyramids are combined through element-wise summation at each pyramid level and fed into the Trident localization head.

\paragraph{Feature selection and gating.} Several works~\cite{hu2018squeeze,woo2018cbam} demonstrate the importance of feature selection in improving the performance of deep learning models. Squeeze-and-excitation networks~\cite{hu2018squeeze} improve the quality of features extracted by a network by explicitly modeling the dependencies between convolutional feature channels. The proposed block initially squeezes global spatial information into a channel descriptor and then excites the descriptor to selectively emphasize important features while suppressing less useful ones. Convolutional Block Attention Module~(CBAM)~\cite{woo2018cbam} enhances feature representation by sequentially inferring attention maps across the channel and spatial dimensions. These maps are multiplied with the input feature map for adaptive refinement. The Feature Selection and Enhancement architecture~(FEASE)~\cite{zhou2024fease} relies on feature selection and enhancement mechanisms to improve the performance of action recognition models. The network employs multi-scale structures and attention to dynamically prioritize and weigh features.

Feature selection mechanisms have also been applied in the field of large language models. In Memorizing Transformers~\cite{wu2022memorizing}, a learned gating mechanism is incorporated to dynamically manage the balance between short-term and long-term memory focus by adjusting the influence of past and present information.


\section{Proposed Method}

Given a set of untrimmed videos $V = \{ v_i\}_{i=1}^{N}, v_i \in \mathbb{R}^{H \times W \times 3}$ and a set of annotations for the action segments $Y_i = \{(s_a, e_a, c_a)\}_{a=1}^{A_i}$, where each tuple $(s_a, e_a, c_a)$ specifies the start time $s_a$, end time $e_a$, and class $c_a$ for the $a^{th}$ action segment, the goal of TAD is to identify all the action segments $Y_i$ within each input video $v_i$. Similar to other works in the literature, the proposed model operates on temporal visual features obtained from spatio-temporal features extractors, such as I3D~\cite{carreira2017quo} or TSP R(2+1)D~\cite{alwassel2021tsp}. Each video $v_i$ is divided into $T$ segments, each segment consisting of $\Delta$ frames; these segments are processed by the feature extractor to obtain segment features $x_i^t \in \mathbb{R}^{D \times 1}$, where $D$ is the dimensionality of the features. All the segment features are stacked across the temporal axis to obtain a comprehensive temporal visual representation $X = \{x_i\}_{i=1}^N, x_i \in \mathbb{R}^{D \times T}$ for each input video $v_i$. 

\subsection{Method Overview}

\begin{figure}[ht]
\centering
\begin{subfigure}{0.57\linewidth}
\centering
\includegraphics[width=\textwidth]{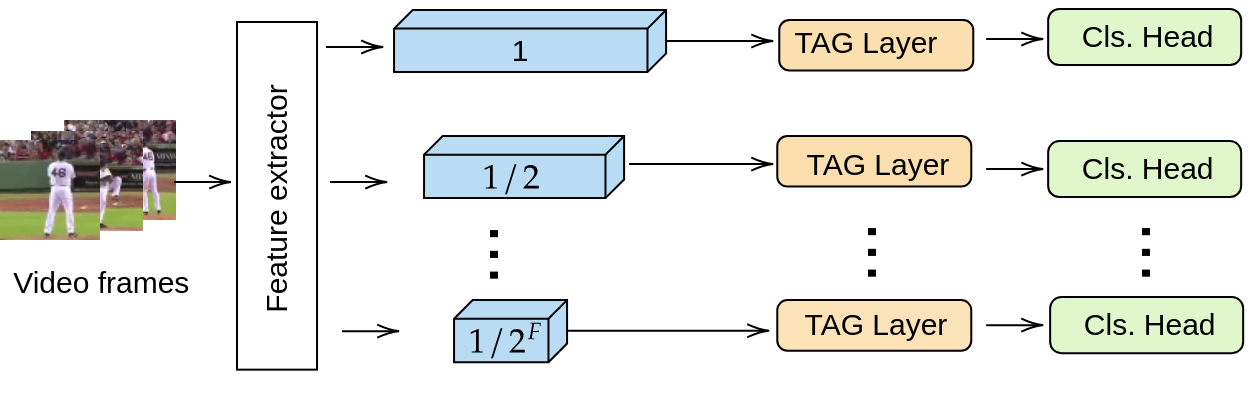}
\caption{Overview Model}
\label{fig:short-a}
\end{subfigure}
\hfill
\begin{subfigure}{0.41\linewidth}
\centering
\includegraphics[width=\textwidth]{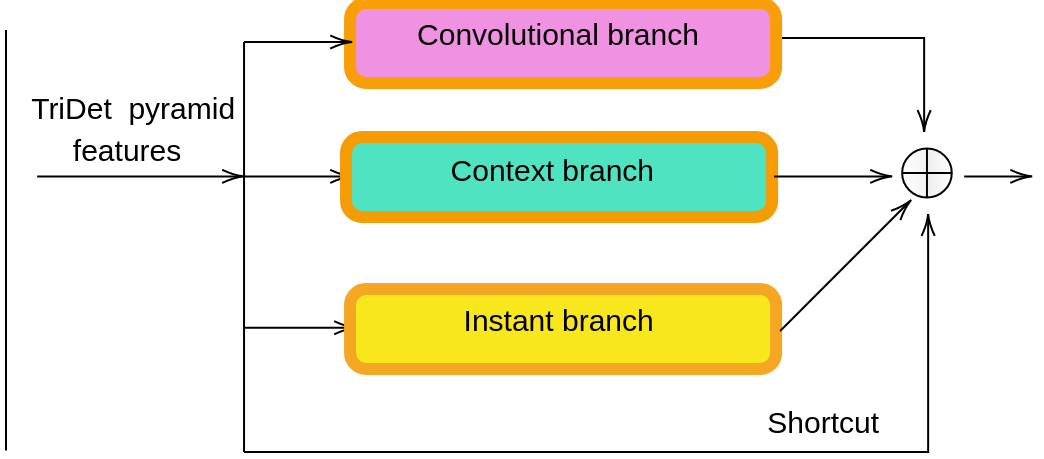}
\caption{Proposed TAG Layer}
\label{fig:short-b}
\end{subfigure}
\caption{Overview of the proposed method. \textbf{(a)}: Based on TriDet, the model consists of a video feature extractor, a feature pyramid extractor, and a boundary-oriented head for action localization and classification. \textbf{(b)}: Structure of the proposed Temporal Attention Gating layer. }
\label{fig:overview}
\end{figure}

This work introduces a one-stage TAD model, built on top of the TriDet architecture~\cite{shi2023tridet}. As its predecessor, the model comprises three modules: a video feature extractor, a feature pyramid extractor that progressively down-samples the video features to effectively handle actions of different lengths, and a boundary-oriented Trident-head~\cite{shi2023tridet} for action localization and classification. The overall architecture is depicted in~\cref{fig:overview}.

The contribution of this work is the design of Temporal Attention Gating~(TAG) in the feature pyramid extractor, as an extension of the Scalable-Granularity Perception~(SGP) layer from the TriDet~\cite{shi2023tridet}. The SGP layer comprises two branches: instant-level and window-level. The instant level branch focuses on distinguishing between action and non-action frames by increasing the distance between the action frames and the average feature of the video. The window-level branch aims to increase the model's ability to capture features at different temporal scales by applying parallel convolutional operations with varying receptive fields. It combines the outputs through summation and then applies weighting to the result.

As an improvement, our TAG layer integrates three branches: context, convolution, and instant. The context branch focuses on the relationship between the central frame and the temporal boundaries of the convolutional operation and allows the model to better capture the temporal dependencies. The convolution branch employs 1D convolutions to extract temporal features and dynamically selects the most informative ones through gating. As in TriDet~\cite{shi2023tridet}, the instant branch focuses on increasing the feature distance between action and non-action frames. The proposed layer provides a more comprehensive and holistic feature representation by combining the strengths of the three branches.

The TAG layer is defined as follows: 

\begin{equation}
  TAG(x) = \Gamma(x) + \Lambda(x) + \Xi(x) + x.
\end{equation}

It adds up the context branch $\Gamma$, the convolution branch $\Lambda$ and the instant branch $\Xi$. The instant branch $\Xi(x) = ReLU(FC(AvgPool(x)))$~\cite{shi2023tridet} uses the temporal average pooling operation $AvgPool$, a Fully Connected~($FC$) layer, and the Rectified Linear Unit~($ReLU$) activation. Context and Convolution branch are described in detail in the next sections.

\subsection{Convolution Branch}

Similar to~\cite{shi2023tridet}, the local branch applies two parallel 1-D convolutional layers $Conv_w$ and $Conv_{kw}$ over the temporal dimension with windows sizes $w$ and $kw$, respectively. This approach allows the model to capture temporal features at multiple scales, providing a richer visual representation. Outputs of these convolutions are concatenated and passed through a Multi-Layer Perceptron~(MLP) to learn a gate $\beta$:
\begin{equation}
\beta (x_i)= \sigma(\mathrm{MLP}^g(Conv_w(x_i) || Conv_{kw}(x_i)) ),
\label{eq:beta1}
\end{equation}
where $||$ is the concatenation operator, $\sigma$ is the sigmoid activation function, and $k$ is a hyperparameter for the convolution window size. For the $w$ and $k$ hyper-parameters we use the same values as in ~\cite{shi2023tridet}.

The gating mechanism balances contributions from the two temporal scales by generating a weighting coefficient, $\beta$. A linear combination of the outputs from different convolutional layers allows the model to compute local features. 
Formally, the local branch $\Lambda$ is defined as:
\begin{equation}
\Lambda(x_i) = \beta \cdot Conv_w(x_i) + (1 - \beta) \cdot Conv_{kw}(x_i).
\label{eq}
\end{equation}

The final local features incorporate both fine-grained and coarse-grained temporal information. By adaptively weighting the contributions from the convolutional layers, the model can prioritize the most relevant temporal patterns in video.

\begin{figure}[ht]
\centering
\includegraphics[width=0.75\linewidth]{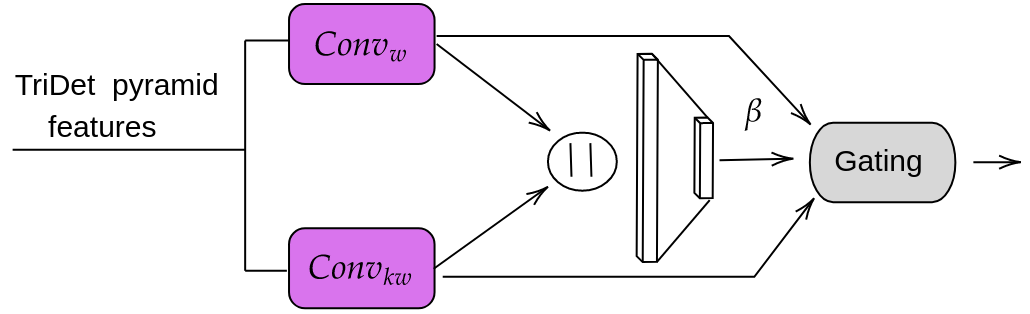}
\caption{\textbf{Convolution branch}: The video features are processed by two parallel convolutions, $Conv_w$ and $Conv_{kw}$ with different temporal sizes. Their responses are concatenated and then passed through a gating mechanism, which predicts a scalar parameter $\beta$ used to combine the features through linear interpolation.}
\label{fig:gating}
\end{figure}

\subsection{Context Branch}

Inspired by the PDAN architecture~\cite{dai2021pdan}, we introduce a new context branch in the feature extractor to enhance temporal action detection by capturing the contextual relationships between frames. 

This branch begins by extracting the boundary frames located at the leftmost and rightmost positions within the receptive field of the convolutional filter used in the local branch. Serving as key-values in the cross-attention mechanism, these frames encapsulate temporal limits and can provide contextual information about the action sequence. The central frame of the convolutional receptive field is used as a query in the cross-attention operation. This frame is interpreted as the center point of the action and serves as the focal point around which the context information is aggregated. 

The attention mechanism in this branch involves interaction between the central frame~(query) and boundary frames
\begin{equation}
Attn(Q,K,V) = softmax\left(\frac{Q \cdot K^T}{\sqrt D}\right)\cdot V
\label{eq:attention}
\end{equation}
where $Q$ is the central frame feature representation, $K$ and $V$ represent the keys and values, and $D$ is the feature dimensionality. Through this branch, the feature representation of the central frame is improved by incorporating relevant information from the boundary frames, leading to a more context aware representation.

\begin{figure}[ht]
\centering
\includegraphics[width=0.75\linewidth]{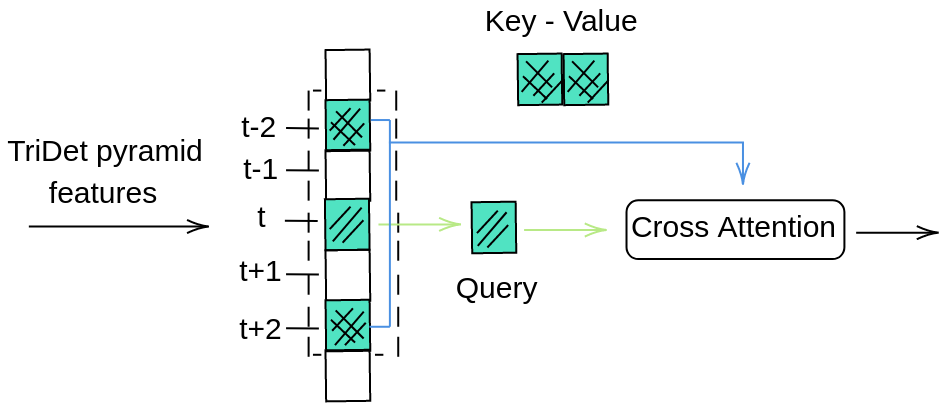}
\caption{\textbf{The context branch} employs cross-attention to include context in the convolution's central frame~(query) representation using boundary frames~(key-values), $t$ is the frame index.}
\label{fig:context}
\end{figure}

\subsection{Prediction and Training Setup}

We adopt the prediction and training framework of TriDet~\cite{shi2023tridet}. The prediction head employs a statistical boundary localization method, which consists of three parts: start head, end head, and center-offset head, respectively responsible for detecting the start, end, and temporal center of action. Additionally, the prediction head uses multiple bins to improve the accuracy of boundary predictions. To minimize parameters, the prediction head is shared across all layers of the feature pyramid.

Each layer $l$ of the feature pyramid generates a temporal feature $F^l \in \mathbb{R}^{(2^{l-1}T) \times D}$, which is used for classification and action instance detection. The output for each instant $t$ in layer $l$ is $\hat{y}^l_t = (\hat{c}^l_t, \hat{d}^l_{\text{st}}, \hat{d}^l_{\text{et}})$, where $\hat{c}^l_t$ is the classification score, and $\hat{d}^l_{\text{st}}$ and $\hat{d}^l_{\text{et}}$ denote the predicted start and end distances, respectively.

The loss function is defined as:
\begin{equation}
\mathcal{L} = \frac{1}{N_{\text{pos}}} \sum_{l,t} \mathds{1}_{\{c_{t}^{l} > 0\}} (\sigma_{\text{IoU}} \mathcal{L}_{\text{cls}} + \mathcal{L}_{\text{reg}}) +
\frac{1} {N_{\text{neg}}} \sum_{l,t} \mathds{1}_{\{c_{t}^{l} = 0\}} \mathcal{L}_{\text{cls}},
\end{equation}
$\mathcal{L}_{\text{cls}}$ and $\mathcal{L}_{\text{reg}}$ represent the classification~(focal) loss~\cite{lin2017focal} and the regression~(IoU) loss~\cite{rezatofighi2019generalized}, respectively. $\sigma_{\text{IoU}}$ is the temporal IoU between the predicted and ground truth segments, and it weights the classification loss to emphasize high-quality regression instances. $N_{\text{pos}}$ and $N_{\text{neg}}$ are the counts of positive and negative samples. $\sigma_{\text{IoU}}$ weights the classification loss to emphasize high-quality regression instances. The classification loss $\mathcal{L}_{\text{cls}}$ is applied on all instances, while the regression loss is applied only to positive samples~(where $c_{t}^{l} > 0$). Positive samples are selected via center sampling, focusing on instances near the action center~\cite{tian2022fully, zhang2022actionformer}. During inference, instances with classification scores above a threshold $\lambda$ are retained, and Soft-NMS~\cite{bodla2017soft} is applied for the deduplication of instances.

\section{Experiments and Results}

All the experiments~(both training and evaluation) were performed on a single NVIDIA A40 GPU using CUDA version $11.2$ and torch version $1.11.0$. We conducted experiments on two action localization benchmarks: THUMOS14~\cite{idrees2017thumos} and EPIC-KITCHENS 100~\cite{damen2022rescaling}. Following established practices~\cite{cheng2022tallformer, lin2019bmn, zeng2019graph, zhang2022actionformer, shi2023tridet}, we use the validation set and report mean average precision~(mAP) at various intersections over union~(IoU) thresholds. 

\paragraph{THUMOS14} includes YouTube videos categorized into $20$ different sports action classes. It comprises $200$ training videos, which encompass $3,007$ action instances, and $213$ validation videos, containing $3,358$ action instances. 
We use I3D~\cite{carreira2017quo} as a backbone feature extractor. The initial learning rate is set to $10^{-4}$ and it is updated using Cosine Annealing scheduler~\cite{loshchilov2016sgdr} for $40$ epochs, of which $20$ are warmup epochs.

\cref{tab:performance_thumos} presents the comparison of the proposed method with several state-of-the-art approaches on the THUMOS14 dataset. The performance is evaluated using mAP across IoU thresholds ranging from $0.3$ to $0.7$. The proposed method achieves the best average performance~($69.1$) across all evaluated IoU thresholds, indicating its effectiveness in detecting temporal actions accurately. Notably, our method outperforms the state-of-the-art baseline TriDet~\cite{shi2023tridet} at higher thresholds~($0.4$, $0.5$, $0.6$, $0.7$) and shows competitive performance at the lowest threshold~($0.3$). Specifically, the proposed method achieves $83.5$, slightly lower than TriDet’s $83.7$. This slight drop at the lowest threshold could be explained model's focus on higher precision detections, which can sometimes result in minor trade-offs at lower thresholds. Only Trident was implemented and evaluated. The other results were reported from their respective papers.

\begin{table}[ht]
\caption{Results on THUMOS14 Dataset. Avg column represents the average mAP across all thresholds. The performance is evaluated using mAP across IoU thresholds ranging from $0.3$ to $0.7$.}
\centering
\begin{tabular}{l|rrrrrr}
\toprule
\textbf{Method} & \textbf{0.3} & \textbf{0.4} & \textbf{0.5} & \textbf{0.6} & \textbf{0.7} & \textbf{Avg} \\
\midrule
BMN~\cite{lin2019bmn} & 56.0 & 47.4 & 38.8 & 29.7 & 20.5 & 38.5 \\
G-TAD~\cite{xu2020g} & 54.5 & 47.6 & 40.3 & 30.8 & 23.4 & 39.3 \\
TCANet~\cite{qing2021temporal} & 60.6 & 53.2 & 44.6 & 36.8 & 26.7 & 44.3 \\
ContextLoc~\cite{zhu2021enriching} & 68.3 & 63.8 & 54.3 & 41.8 & 26.2 & 50.9 \\
ReAct~\cite{shi2022react} & 69.2 & 65.0 & 57.1 & 47.8 & 35.6 & 55.0 \\
TadTR~\cite{liu2022end} & 74.8 & 69.1 & 60.1 & 46.6 & 32.8 & 56.7 \\
TALLFormer~\cite{cheng2022tallformer} & 76.0 & - & 63.2 & - & 34.5 & 57.9 \\
ActFormer~\cite{zhang2022actionformer} & 82.1 & 77.8 & 71.0 & 59.4 & 43.9 & 66.8 \\
TriDet\tablefootnote{retrained method}~\cite{shi2023tridet}& \textbf{83.7} & 79.5 & 72.2 & 61.4 & 45.8 & 68.5 \\ \hline
\textbf{Ours} & 83.5 & \textbf{79.6} & \textbf{72.9} & \textbf{61.9} & \textbf{47.5} & \textbf{69.1} \\
\bottomrule
\end{tabular}
\label{tab:performance_thumos}
\end{table}

\paragraph{EPIC-KITCHENS 100} is a large-scale first-person vision dataset focusing on two tasks: noun localization, i.e. identifying objects like doors, and verb localization, i.e. identifying actions. For this dataset, we use SlowFast~\cite{feichtenhofer2019slowfast} as the backbone feature extractor. For both noun and verb subsets, the network is trained with an initial learning rate of $10^{-4}$ updated with the Cosine Annealing scheduler~\cite{loshchilov2016sgdr}. For noun localization, we trained for $29$ epochs, of which $5$ are warmup epochs, while for verb localization, we trained for $27$ epochs, of which $5$ are warmup epochs.

\cref{tab:performance_vn} reports the mAP at thresholds ranging from $0.1$ to $0.5$ on the EPIC-KITCHEN 100 dataset. The results illustrate the superior performance of the proposed method across all evaluated thresholds. For verb detection, our method achieved an mAP of $28.7\%$ for IoU threshold $0.1$, demonstrating substantial improvement over the next best model, TriDet, which scored $27.8\%$. In the noun detection category, our method similarly outperforms existing approaches, starting at $27.4\%$ for IoU $0.1$ and reaching $18.2\%$ at IoU threshold $0.5$. The average across thresholds is $23.6\%$ and it indicates the ability of the model to recognize relevant objects involved in the actions.

\begin{table}[ht]
\caption{Results on EPIC-KITCHEN 100 dataset for the Verb and Noun sub-tasks. Avg column represents the average mAP across all thresholds. The performance is evaluated using mAP across IoU thresholds ranging from $0.1$ to $0.5$.}
\centering
\begin{tabular}{l|rrrrrr|rrrrrr}
\toprule
& \multicolumn{6}{c|}{\textbf{Verb}} & \multicolumn{6}{c}{\textbf{Noun}} \\
\midrule
\textbf{Method} & \textbf{0.1} & \textbf{0.2} & \textbf{0.3} & \textbf{0.4} & \textbf{0.5} & \textbf{Avg} & \textbf{0.1} & \textbf{0.2} & \textbf{0.3} & \textbf{0.4} & \textbf{0.5} & \textbf{Avg} \\
\midrule
BMN~\cite{lin2019bmn} & 10.8 & 8.8 & 8.4 & 7.1 & 5.6 & 8.1 & 10.3 & 8.3 & 6.2 & 4.5 & 3.4 & 6.5\\
G-TAD~\cite{xu2020g} & 12.1 & 11.0 & 9.4 & 8.1 & 6.5 & 9.4 & 11.0 & 10.0 & 8.6 & 7.0 & 5.4 & 8.4 \\
ActFormer~\cite{zhang2022actionformer} & 26.6 & 25.4 & 24.2 & 22.3 & 19.1 & 23.5 & 25.2 & 24.1 & 22.7 & 20.5 & 17.0 & 21.9\\
TriDet\tablefootnote{retrained method}~\cite{shi2023tridet} & 27.8 & 26.9 & 25.4 & 23.2 & 19.5 & 24.5 & 25.9 & 24.7 & 22.9 & 20.7 & 17.4 & 22.3 \\
\midrule
 \textbf{Ours} & \textbf{28.7} & \textbf{27.7} & \textbf{26.1} & \textbf{23.5} & \textbf{20.4} & \textbf{25.3} & \textbf{27.4} & \textbf{26.4} & \textbf{24.1} & \textbf{22.0} & \textbf{18.2} & \textbf{23.6}\\
\bottomrule
\end{tabular}
\label{tab:performance_vn}
\end{table}


\section{Ablation Studies}

The ablation studies are performed on the THUMOS14 dataset~\cite{idrees2017thumos} and the performance is measured using the mAP at different intersections over union~(IoU) thresholds~($0.3$, $0.4$, $0.5$, $0.6$, $0.7$) similar as in the experimental results section. The \textit{Avg} column represents the average mAP across all thresholds.

\subsection{Layer Components}

\cref{tab:ablation_components} presents an ablation study to evaluate the influence of the context~(cross attention) and gating mechanisms on the performance of the proposed layer.

\begin{table}[ht]
\caption{Local and global branch influence. The performance is evaluated on THUMOS14
dataset using mAP across IoU thresholds ranging from $0.3$ to $0.7$. }
\centering
\begin{tabular}{l|rrrrrr}
\toprule
\textbf{Setup} & \textbf{0.3} & \textbf{0.4} &\textbf{0.5} & \textbf{0.6} & \textbf{0.7} & \textbf{Avg} \\
\midrule
without gating, without context & \textbf{83.7} & 79.5 & 72.2 & 61.4 & 45.8 & 68.5 \\
with gating, without context & 83.3 & 79.8 & 72.8 & 61.7 & 46.6 & 68.8 \\
without gating, with context & 83.6 & 79.2 & 72.7 & 61.6 & 46.7 & 68.8 \\ \hline
with gating, with context  & 83.5 & \textbf{79.6} & \textbf{72.9} & \textbf{61.9} & \textbf{47.5} & \textbf{69.1}\\ 
\bottomrule
\end{tabular}
\label{tab:ablation_components}
\end{table}

Excluding the context branch leads to a decrease in performance, particularly at higher thresholds. The lack of cross attention likely reduces the model's ability to capture relationships between the extreme frames and the central frame, leading to less precise feature selection and localization. On the other hand, by removing the gating mechanism a slightly lower average mAP is achieved than compared to including both mechanisms. While the context branch still provides some performance improvement, the lack of gating means that the model cannot effectively prioritize the most relevant features, leading to a lower performance. Finally, including both the gating mechanism and the context branch yields the highest performance across the average mAP and on all thresholds except the smallest one, where without gating and context is more beneficial. While the gating mechanism is effective at selecting the most relevant features, it might introduce some overhead or complexity that is not strictly necessary at lower thresholds. 

This experiment shows that the gating mechanism enhances feature selection by prioritizing the most relevant features, while the context branch~(cross-attention) improves the model's ability to understand the relationships between frames. This combination results in better localization and higher overall accuracy, as shown by the highest average mAP when both branches are used.

\subsection{Gating Mechanism}

In this section, we examine the effectiveness of the gating mechanism. The proposed TAG layer utilizes two parallel convolutions~($Conv_w$ and $Conv_{kw}$) and employs a gating mechanism to select the most relevant features. \cref{tab:ablation_gating} presents a comparison between the gating mechanism and other strategies for fusing the convolution responses. In the table, "Average" denotes the strategy of simply averaging the feature maps from the two convolutions:
\begin{equation}
Average(x) = \frac{Conv_w(x) + Conv_{kw}(x)}{2}.
\end{equation}
"Maximum" involves taking the maximum value across each feature:
\begin{equation}
Maximum(x) = \max(Conv_w(x), Conv_{kw}(x)).
\end{equation}
Finally, "Baseline" refers to the approach described in~\cite{shi2023tridet}, where the responses of the two convolutions are summed and then weighted by the response of another convolution:
\begin{equation}
Baseline(x) = Conv_{w'}(x) \times (Conv_w(x) + Conv_{kw}(x)).
\end{equation}

\begin{table}[ht]
\caption{Ablation study on the influence of the gating mechanism for the THUMOS14 dataset. The performance is evaluated using mAP across IoU thresholds ranging from $0.3$ to $0.7$.}
\centering
\begin{tabular}{l|rrrrrr}
\toprule
\textbf{Setup} & \textbf{0.3}& \textbf{0.4} & \textbf{0.5} & \textbf{0.6} & \textbf{0.7} & \textbf{Avg} \\
\midrule
Baseline~\cite{shi2023tridet} & \textbf{83.7} & 79.5 & 72.2 & 61.4 & 45.8 & 68.5 \\
Average & 83.1 & 79.2 & 71.8 & 61.6 & 46.7 & 68.5 \\ 
Maximum & 83.5 & 79.3 & 72.7 & 60.7 & 46.2 & 68.5\\ \hline
Gating~(Ours) & 83.5 & \textbf{79.6} & \textbf{72.9} & \textbf{61.9} & \textbf{47.5} & \textbf{69.1}\\ 
\bottomrule
\end{tabular}
\label{tab:ablation_gating}
\end{table}

The proposed method surpasses all other methods on the average mAP and across the higher thresholds, especially notable at the highest threshold~($0.7$) with a score of $47.5$. On the lowest threshold~($0.3$), the baseline method achieves the best performance. Lower IoU thresholds allow for a more relaxed strategy in matching predicted bounding boxes with ground truth boxes; as the threshold increases, the criteria for a successful match become stricter, requiring more precise localization of predicted boxes. The Baseline method seems to favor lower thresholds, where approximate matches are sufficient. By simply summing and weighting the convolution responses, the \textit{Baseline} method may lose detailed spatial information that is relevant for higher IoU thresholds. This loss can lead to inaccuracies in object localization, thereby reducing mAP at stricter thresholds.

The results indicate that the gating mechanism effectively enhances feature selection, leading to better overall performance compared to averaging, taking the maximum, or the baseline approach. The gating mechanism achieves the highest average mAP, demonstrating its ability to select relevant features and improve model performance across different threshold levels.

\section{Conclusions and Future Work}

In this study, we introduced a new TAG layer in the feature pyramid extractor of TAD models, which includes convolutional operations paired with a gating mechanism, and a context-aware module based on cross-attention. The TAG layer incorporates three branches: a convolution branch, a context branch, and an instant branch. The local branch uses two convolutions with different window sizes in parallel, to extract both fine-grained and coarse-grained temporal features. The integration of a gating mechanism further refines the feature selection process, ensuring that only the most relevant features are processed for action detection. 
The context branch employs the outermost frames from the convolution operation as key-value pairs and the central frame as the query, allowing the model to understand the temporal context of the action sequence.

Conducted on two TAD benchmarks, EPIC-KITCHEN 100 and THUMOS14, our experiments confirm the efficacy of our proposed model. The results demonstrate an improvement in detection performance compared to existing methods, underlining the benefits of the proposed architectural design.

In future work, we aim to extend the applicability of our method across diverse model architectures to enhance its generalizability. Integrating multimodal data, such as audio and text annotations, could further refine action detection capabilities. 


%
%
\bibliographystyle{splncs04}
\bibliography{bibliography}
\end{document}